\documentclass[a4paper,fleqn]{cas-sc}

\usepackage[authoryear,longnamesfirst]{natbib}
\usepackage{listings}   
\usepackage{tabularx}   
\usepackage[section]{placeins}

\begin{document}

\shorttitle{Expression of self-stigma in online substance use communities}
\shortauthors{Bouzoubaa et al.}

\title[mode=title]{The cognitive, affective, and behavioral expression
of self-stigma among people who use drugs in online substance use
communities}

\author[1]{Layla Bouzoubaa}
\cormark[1]
\ead{lb3338@drexel.edu}

\author[1]{Hyung Wook Choi}
\author[1]{Milan Varghese}
\author[2]{Valerie Earnshaw}
\author[1]{Rezvaneh Rezapour}

\affiliation[1]{organization={Department of Information Science, Drexel University},
            state={Pennsylvania}, country={USA}}
\affiliation[2]{organization={Department of Human Development and Family Sciences,
            University of Delaware}, state={Delaware}, country={USA}}

\cortext[1]{Corresponding author}

\begin{abstract}
\textbf{Objectives:} To develop a codebook for self-stigma across cognitive, affective, and behavioral domains, and to estimate the prevalence, co-occurrence, and temporal patterns of these indicators in Reddit posts by people who use drugs.\\
\textbf{Methods:} First, we developed a ten-indicator codebook through consensus-based abductive coding of Reddit posts, spanning cognitive (self-labeling, pessimism/self-defeatism, deservingness/worthlessness), affective (shame, guilt/self-blame, despair/hopelessness), and behavioral (concealment, anticipated rejection, desire to quit, ambivalence) domains; two trained coders applied the final codebook with substantial agreement (Cohen's $\kappa = 0.723$). Second, we applied the codebook to examine indicator co-occurrence and emergence across users' posting timelines, scaling classification with a large language model validated against expert human coding ($\kappa = 0.730$, $F_1 = 0.802$). The applied analysis used 72,115 thread-initiating posts from 1,660 English-language Reddit users with sustained engagement in substance use communities (2006-2025).\\
\textbf{Results:} Of 72,115 posts, 3,838 (5.3\%) contained self-stigma expressions from 1,228 users (74.0\%); all ten indicators discriminated self-stigma from non-self-stigma posts (RR 3.6 to 86.2), with self-labeling (56.0\%) and despair/hopelessness (48.5\%) most prevalent. Self-stigma was expressed as an integrated phenomenon: core and behavioral indicators were strongly associated at the user level (OR = 4.65, 95\% CI 3.12--6.94, $p < 0.001$), and 87.0\% of posts containing behavioral indicators also contained a core indicator. Contrary to progressive models, behavioral indicators emerged earlier than core indicators in users' posting sequences (desire to quit at median position 0.08 vs. shame at 0.38). Nine of ten indicators were stable across posting trajectories; only pessimism/self-defeatism increased (OR = 1.62, 95\% CI 1.25--2.10, FDR-adjusted $p = 0.002$).\\
\textbf{Conclusion:} Among people who use drugs engaging in online communities, self-stigma is expressed as an integrated phenomenon where behavioral indicators rarely appear without internalized ones and frequently precede cognitive and affective expressions. While most self-stigma expressions remain stable over time, pessimism regarding the possibility of change deepens, highlighting a specific target for early digital intervention and demonstrating that progressive stage models of self-stigma development do not translate directly to textual disclosure.
\end{abstract}


\maketitle

\section*{Introduction}\label{intro}

In the United States, 80\% of the 48 million people who meet criteria for substance use disorder (SUD) do not receive treatment \cite{SAMHSA2024}. Structural barriers such as cost and availability contribute to this gap, but they cannot fully explain why people who use drugs (PWUD) avoid seeking help even when services are accessible. A growing body of evidence points to stigma, particularly internalized stigma, as a fundamental obstacle that operates at the psychological core of treatment avoidance \citep{hammarlundReviewEffectsSelfstigma2018, hatzenbuehlerStigmaFundamentalCause2013}. Because self-stigma is theorized to develop progressively \citep{corriganSelfStigmaMentalIllness2012}, the timing at which its components emerge may itself matter for understanding how and when to intervene. Yet how PWUD spontaneously articulate self-stigma, and how its components co-occur and unfold over time, remains underexplored, in part because the vast majority of self-stigma research has been confined to cross-sectional surveys of treatment-seeking samples \citep{spataSubstanceUseStigma2024, earnshawAdvancingSubstanceUse2024}.


Public attitudes toward PWUD are significantly more negative than toward people with other health conditions: only 22\% of Americans are willing to work closely with someone with SUD compared to 62\% for mental illness, and 90\% would not want someone with SUD to marry into their family \citep{barryPublicStigma2014}. This pronounced social rejection creates the environment within which self-stigma develops. Self-stigma is conceptualized as the endorsement of negative beliefs and feelings associated with one's stigmatized status, distinct from but related to enacted stigma (past discrimination) and anticipated stigma (expectations of future mistreatment) \cite{earnshawStigmaSubstanceUse2020}. Theory distinguishes core components of self-stigma, cognitive endorsement of stigmatized identity, and affective responses such as shame \citep{luomaSelfStigmaSubstanceAbuse2013}, from behavioral consequences such as concealment and disengagement from valued life goals \citep{corriganSelfStigmaMentalIllness2012}.
Corrigan's progressive model \citep{corrigan2011examining} describes self-stigma as developing through stages of awareness, agreement, and self-application, culminating in the ``why try'' effect: a diminished sense that pursuing recovery is worthwhile \citep{corrigan2011examining, corrigan2009whytry}. Self-stigma predicts treatment avoidance, early dropout, and poorer outcomes \citep{livingstonCorrelatesConsequencesInternalized2010}.

Self-stigma is identified through the indicators it produces in language and behavior: how individuals describe themselves, what they feel, and how they act in response to their stigmatized status. Existing research has operationalized these indicators primarily through structured scales such as the Substance Abuse Self-Stigma Scale \citep[(SASSS);][]{luomaSelfStigmaSubstanceAbuse2013}, the Substance Use Stigma Mechanisms Scale \citep[(SU-SMS);][]{smithSubstanceUseStigma2016}, and the Brief Opioid Stigma Scale \citep[(BOSS);][]{yangBriefOpioidStigma2019}. These instruments have advanced understanding considerably, but they share three limitations relevant to the present study. First, nearly all validation studies use cross-sectional designs, and few measures demonstrate test-retest reliability \citep{spataSubstanceUseStigma2024}. Second, validation samples have been predominantly White and treatment-seeking, restricting what can be said about PWUD outside formal care \citep{spataSubstanceUseStigma2024}. Third, predetermined response categories cannot capture how self-stigma manifests in the spontaneous language of individuals describing their own experiences. Evidence that social desirability bias substantially affects self-reported stigma \citep{latkinSocialDesirabilityResponse2017}, and that behavioral indicators of shame predict relapse more strongly than self-report measures \citep{randlesTracy2013nonverbal}, suggests that naturalistic observation may capture dimensions of self-stigma that survey measures miss.


Social media platforms, particularly Reddit, offer a direct methodological counter to these survey limitations. Reddit's pseudonymity removes the social desirability pressures inherent in clinical reporting, reducing barriers to the spontaneous disclosure of stigmatized behaviors \citep{bouzoubaa_decoding_2024, bouzoubaa_stigma_2024, bouzoubaaPhenotypesStigmaExpressed2026}. Users' posting histories provide records spanning months or years of sustained engagement with substance use communities \citep{macleanForum77AnalysisOnline2015, luInvestigateTransitionsDrug2019, Bouzoubaa_Young_Rezapour_2024}. These platforms reach individuals not represented in treatment-based samples, including those who have not engaged with formal care \citep{SAMHSA2024}. Prior computational work has shown that natural language processing (NLP) approaches can identify substance use stigma in social media text \citep{chenExaminingStigmaRelating2022, roeslerTheoryinformedDeepLearning2024, bouzoubaaPhenotypesStigmaExpressed2026}, with self-stigma expressions characterized by shame, self-blame, and despair \cite{roeslerTheoryinformedDeepLearning2024}. Self-stigma has been documented as one of the most frequently exhibited types of stigma across substance use communities \cite{chenExaminingStigmaRelating2022}. However, existing computational work has mostly treated self-stigma as monolithic rather than examining its constituent cognitive, affective, and behavioral indicators, limiting what can be said about how these dimensions co-occur or emerge across users' engagement with these communities \citep{fox2025advancing}.

\subsection*{The present study}

Responding to recent calls for substance use stigma research to incorporate temporal considerations \citep{earnshawAdvancingSubstanceUse2024, fox2025advancing}, this study develops a theory-informed framework for characterizing self-stigma expression in naturalistic online discourse and applies it at scale to examine how its components co-occur and unfold over time. Using a hybrid approach combining consensus-based qualitative coding with validated large language model (LLM) classification, we analyze 72,115 posts from 1,660 Reddit users who maintained sustained engagement with substance use communities over at least 180 days. We pursue three aims: First, we develop a ten-indicator codebook spanning the cognitive, affective, and behavioral domains posited by self-stigma theory \citep{corriganSelfStigmaMentalIllness2012, earnshawStigmaSubstanceUse2020, luomaSelfStigmaSubstanceAbuse2013} to enable granular identification of self-stigma manifestations in text. Second, we aim to apply this framework to characterize indicator prevalence, the co-occurrence of core and behavioral components, and the temporal dynamics of expression across users' posting histories. Third, we validate a scalable, mixed-methods framework, combining qualitative coding with LLMs, to advance the methodological toolkit for identifying self-stigma in naturalistic, longitudinal text. Detecting these dimensions and their dynamics in naturalistic discourse could inform when and how stigma-reduction and engagement efforts reach PWUD outside formal care. 

\textit{Study Design \& Epistemological Approach:} We treat the patterns described here as features of language use in online discourse rather than direct evidence of underlying psychological processes. The emergence of an indicator in a user's posting history reflects when a user first disclosed that experience in text, which may or may not correspond to when they first experienced it.
\section*{Methods}\label{methods}

\subsection*{Data collection}\label{subsec:data}
 
Reddit (\url{https://reddit.com}) is a pseudonymous social media platform organized into topic-based communities called ``subreddits.'' Substance use communities on Reddit are spaces where PWUD share experiences and discuss their substance use openly, often with greater candor than in clinical or offline settings. We analyzed self-stigma expressions across users' complete Reddit posting histories, including posts in both substance use-focused communities (e.g., \textit{r/opiates}, \textit{r/Stims}) and general interest subreddits (e.g., \textit{r/gaming}, \textit{r/AskReddit}).
 
\begin{figure}[!ht]
    \centering
    \includegraphics[width=0.55\linewidth]{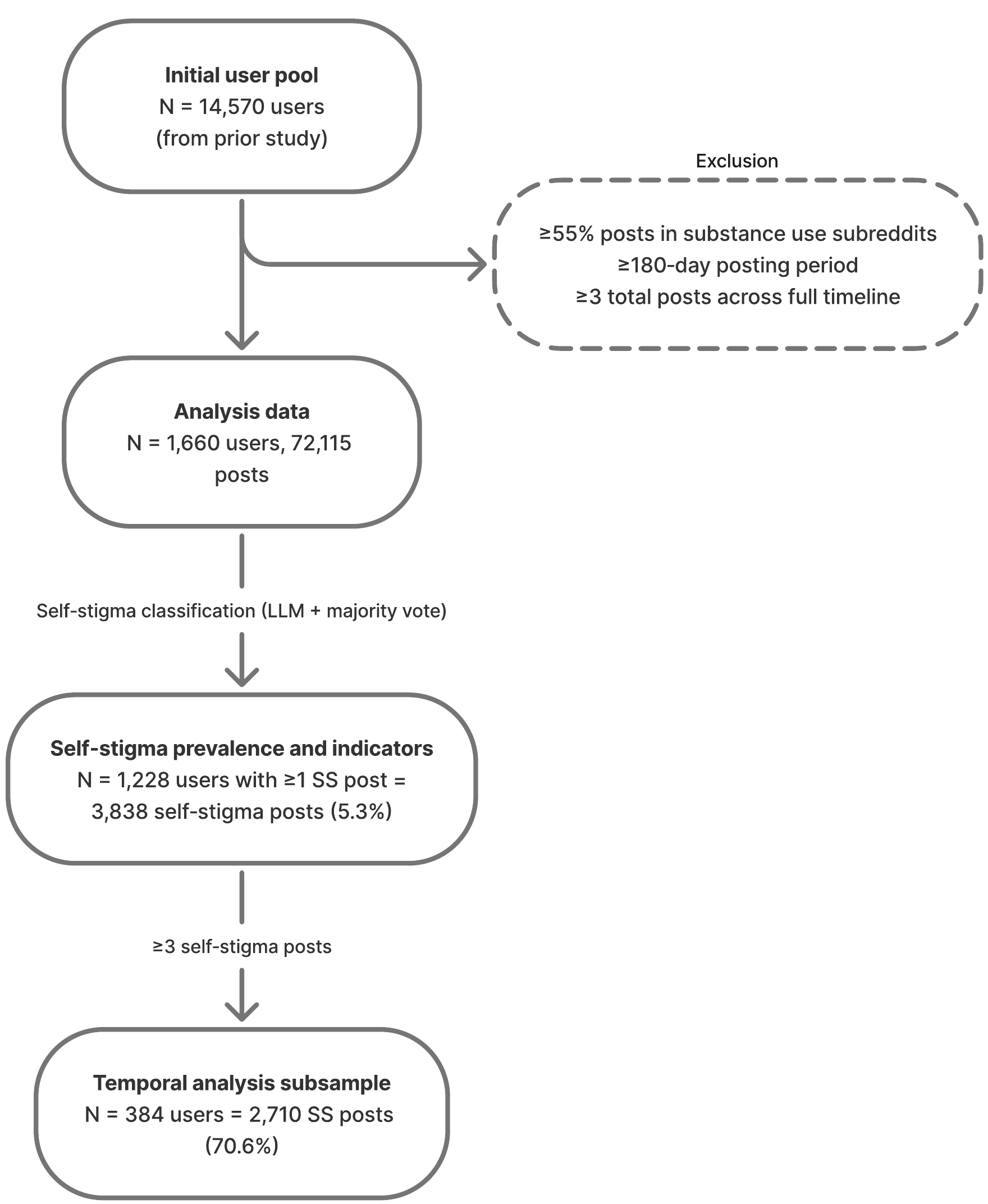}
    \caption{Sample selection and analytic stages. SS = self-stigma.}
    \label{fig:sample_selection}
\end{figure}

Our initial sample of 14,570 English-language Reddit users was drawn from a larger corpus collected for prior work on stigma expression \citep{bouzoubaaPhenotypesStigmaExpressed2026}, in which each user had at least one post classified as containing self-stigma. These users posted in substance use subreddits between March 2006 and August 2025. Our corpus consists exclusively of thread-initiating posts (submissions) rather than comments, as submissions tend to contain more extensive self-disclosure and narrative detail \citep{luInvestigateTransitionsDrug2019, macleanForum77AnalysisOnline2015}. To capture users with sustained engagement rather than casual visitors, we retained users who met three criteria: at least 55\% of posts in substance-use subreddits (the 75th percentile of the initial sample), a minimum 180-day posting span, and at least three total posts. The final sample comprised 1,660 users contributing 72,115 posts across their complete Reddit timelines. Figure~\ref{fig:sample_selection} summarizes the sample selection at different stages of this study. This study analyzed publicly available, pseudonymized data and was deemed exempt from human subjects review by the Institutional Review Board at Drexel University.

\subsection*{Self-stigma classification}\label{subsec:ss_classification}
 
To identify posts containing expressions of self-stigma across a user's complete timeline (i.e., outside of subreddits analyzed in prior work \citep{bouzoubaaPhenotypesStigmaExpressed2026}), we used an iterative prompt engineering approach anchored in a manually coded development set \citep{wang_human_2024, tan-etal-2024-large}. Two annotators independently coded 350 randomly selected posts (150 positive, 200 negative for self-stigma), distinguishing negative self-evaluations or identity-level condemnation related to substance use from situational distress or help-seeking without self-condemnation. We then used this development set to evaluate multiple closed- and open-weight LLMs and selected GPT-4.1-mini (\texttt{gpt-4.1-mini-2025-04-14}; \cite{openaiIntroducingGPT41API2025}) based on agreement with human coders (Cohen's $\kappa = 0.769$, $F_1 = 0.856$ on the development set) and cost-effectiveness (See Appendix \ref{supp:model_comparison} for additional details).
 
Due to the subjective nature of stigma interpretation, we implemented a majority-voting strategy: each post was classified three times by GPT-4.1-mini with identical prompts but independent API calls, and the modal response was taken as the final classification \citep{tan-etal-2024-large, wang_human_2024}. We then validated the stabilized prompt on a stratified random sample of 300 held-out posts (100 self-stigma positive, 200 self-stigma negative) not used during prompt development. The classifier maintained substantial agreement with expert human coding (Cohen's $\kappa = 0.730$, $F_1 = 0.802$; Appendix Table \ref{tab:supp_validation_metrics}). We applied the validated classifier to the entire corpus ($N = 72,115$). 

\FloatBarrier

\subsection*{Self-stigma indicators codebook and classification}
 
Existing measures of self-stigma center on a small set of core features such as shame and avoidant behaviors \citep{luomaSelfStigmaSubstanceAbuse2013}, negative self-labeling, feelings of worthlessness and self-devaluation \citep{corrigan2006blame, corrigan2009whytry}, and discreditation of oneself \citep{smithSubstanceUseStigma2016}. However, prior computational work indicates that stigma expression in naturalistic text encompasses additional content, including despair, motivational concerns, and behavioral responses such as concealment \citep{roeslerTheoryinformedDeepLearning2024, chenExaminingStigmaRelating2022, kepnerTypesSourcesStigma2022}. We therefore developed a codebook to characterize the cognitive, affective, and behavioral content of self-stigma as it appears in text.
 
We used consensus-based abductive coding \citep{timmermans2012theory}, combining deductive codes from the stigma literature with inductive patterns from a purposive sample of Reddit posts varied by substance type and expression frequency (sampling and coding-team details in Appendix \ref{supp:indicator_validation}). The team iteratively refined operational definitions and decision rules through multiple rounds of joint coding and discussion with domain experts in stigma measurement. This process resulted in ten indicators (Table~\ref{tab:codebook}): six \textit{core} indicators capturing the cognitive and affective components of self-stigma \citep{luomaSelfStigmaSubstanceAbuse2013, earnshawStigmaSubstanceUse2020}, and four behavioral indicators capturing the theoretical model's \citep{corrigan2011examining} link to self-stigma. Two trained coders applied the final codebook to 50 randomly sampled self-stigma posts; macro-averaged agreement was substantial (Cohen's $\kappa = 0.723$). Per-indicator agreement and adjudication procedures are reported in Appendix Table \ref{tab:supp_irr_results}.

 
To scale indicator-level coding to the full dataset, we developed one binary classifier per indicator using GPT-4.1-mini (ten in total). Each prompt included operational definitions, decision rules, and exemplar posts. We validated each classifier on 100 held-out self-stigma posts, using stratified sampling to address indicator prevalence variation (full results in Appendix Table \ref{tab:supp_indicator_performance}). We then applied the validated classifiers to all 72,115 posts, enabling indicator detection in posts that did not meet the overall self-stigma threshold. Each post received ten binary indicator labels.

\subsection*{Data analysis}
 
Our analyses proceeded in three stages. First, we examined all 1,660 users to establish self-stigma prevalence and characterize the distribution of expression frequency. Second, we analyzed indicator prevalence and co-occurrence patterns among the users who expressed self-stigma at least once (1,228 users, contributing 3,838 self-stigma posts). Third, we examined temporal patterns among the users with three or more self-stigma posts (384 users, 2,710 posts), enabling analysis of change across users' posting histories.

\paragraph*{Co-occurrence analysis}
 
To examine the theoretical distinction between core and behavioral components of self-stigma, we constructed a 2$\times$2 contingency table cross-classifying users by whether they ever expressed any core indicator (i.e., cognitive or affective) and whether they ever expressed any behavioral indicator across their posting history. We used Fisher's exact test \citep{fisher1922interpretation} to assess the association and reported odds ratios with 95\% confidence intervals. As a complementary descriptive analysis, we examined post-level co-occurrence by calculating the proportion of posts containing behavioral indicators that also contained at least one core indicator. To characterize specific indicator pairings, we calculated phi ($\phi$) coefficients and odds ratios for each of the 45 indicator pairs at the user level, applying Benjamini--Hochberg FDR correction across all pairwise tests \citep{benjamini1995controlling}.

\subsection*{Temporal pattern analysis}\label{subsec:temporal}
 
For temporal analyses, we focused on the subset of users with three or more self-stigma posts ($n$ = 384). For each user, we normalized post order to a 0--1 scale representing position within the user's posting sequence (0 = first post, 1 = last post). We used post-order normalization rather than calendar-based time because posting cadence varied substantially across users. All formal tests of temporal change used generalized estimating equations (GEE) to account for within-user clustering of repeated observations \citep{liang1986longitudinal}, with exchangeable working correlation.
 
\paragraph*{Indicator emergence.} We identified the normalized position at which each indicator first appeared in users' complete posting histories. To test whether core indicators preceded behavioral indicators, we classified users who expressed both indicator types into three categories based on their first-appearance positions: core first, behavioral first, or simultaneous (same post). Users with simultaneous emergence were excluded before testing, and sequencing asymmetry was assessed using an exact two-sided binomial test \citep{clopper1934confidence}.
 
\paragraph*{Stability of expression.} We assessed stability through three complementary approaches. First, we tested whether the breadth of expression (number of co-occurring indicators per self-stigma post) changed across users' trajectories using a GEE model with a Gaussian family and identity link. Second, we tested whether the composition of expression changed by fitting separate binary GEE models (logit link) for each of the ten indicators:
\begin{equation}\label{eq:indicator_trend}
    \text{logit}\bigl(P(\text{Indicator}_{k,ij} = 1)\bigr) = \beta_{0k} + \beta_{1k} \cdot \text{Position}_{ij}
\end{equation}
where $\text{Indicator}_{k,ij}$ denotes the presence (1) or absence (0) of indicator $k$ in post $j$ of user $i$, and $\beta_{1k}$ represents the log-odds change in indicator $k$ prevalence across the posting trajectory. $P$-values were adjusted using Benjamini--Hochberg FDR correction across all ten tests. Third, we compared indicator prevalence in each user's first versus last self-stigma post using McNemar's test for paired binary data \citep{mcnemar1947note}, with FDR correction. Continuous trends were visualized using LOESS with 95\% confidence bands from cluster bootstrap (500 user-level resamples). 

\paragraph*{Declaration of Generative AI}
During the preparation of this work, the authors used OpenAI's GPT4.1-mini (\texttt{gpt-4.1-mini-2025-04-14}) models to assist with data annotation tasks for self-stigma classification and indicator labeling. Additionally, ChatGPT was used to refine the manuscript's readability, voice, and academic tone during the editing process. After using these AI tools, the authors reviewed and edited all content as needed and take full responsibility for the content of the published article.
\section*{Results}\label{results}

 
\subsection*{Sample characteristics}
 
The data comprised 1,660 Reddit users who posted regularly in substance use communities, contributing 72,115 posts over a 19-year period (Table~\ref{tab:sample_characteristics}). Users maintained median posting timelines of approximately two years (726 days, IQR: 347--1,454) and engaged with a median of 5 distinct subreddit communities (IQR: 3--10). Median post length was 61 words (IQR: 19--140).
 
\begin{table}[ht]
\centering
\caption{Sample Characteristics}
\label{tab:sample_characteristics}
\small
\begin{tabular}{lc}
\hline
\textbf{Characteristic} & \textbf{Value} \\
\hline
Users, N & 1,660 \\
Posts, N & 72,115 \\
Posts per user, median (IQR) & 22 (10--47) \\
Timeline days, median (IQR) & 726 (347--1,454) \\
Subreddits per user, median (IQR) & 5 (3--10) \\
Words per post, median (IQR) & 61 (19--140) \\
Date range & Mar 2006 -- Aug 2025 \\
\hline
\end{tabular}
\end{table}
 
The sample was predominantly from opioid-related communities: 35.9\% of posts discussed narcotics, and \textit{r/opiates} alone accounted for 35.0\% of all posts. Other substance categories included general drug discussion (15.6\%), hallucinogens (8.5\%), benzodiazepines (4.6\%), and substitution therapies such as methadone and buprenorphine (2.9\%). Alcohol-related posts were rare (0.6\%), likely reflecting the separation of alcohol-focused communities on Reddit. 

\subsection*{Self-stigma prevalence and distribution}
 
Among 72,115 posts, we identified 3,838 (5.3\%) containing expressions of internalized stigma. Because the analytic sample was drawn from a corpus enriched for stigma expression (\S \ref{subsec:data}), each user in the initial pool had at least one prior stigma-classified post. These estimates describe prevalence within this defined sample rather than base rates among general substance use community participants. The majority of users (74.0\%, $n$ = 1,228) expressed self-stigma in at least one post during their Reddit activity, while 432 users (26.0\%) showed no identifiable self-stigma expression across their posting histories.
 
Self-stigma expression frequency varied substantially across users (Table~\ref{tab:ss_distribution}). Approximately half (50.8\%, $n$ = 844) expressed self-stigma episodically (1--2 posts), while 18.0\% ($n$ = 299) expressed self-stigma in 3--7 posts, and 5.1\% ($n$ = 85) expressed self-stigma in 8 or more posts. Among the high-frequency group, the number of self-stigma posts ranged from 8 to 106 (median = 11), suggesting persistent patterns of self-stigma expression for a subset of users.
 
\begin{table*}[ht]
\centering
\caption{Self-Stigma Expression by User Group}
\label{tab:ss_distribution}
\begin{tabularx}{\textwidth}{lccccc}
\hline
 & \textbf{None} & \textbf{Low} & \textbf{Moderate} & \textbf{High} & \textbf{Total} \\
 & (0 SS posts) & (1--2 SS posts) & (3--7 SS posts) & (8+ SS posts) & \\
\hline
N users (\%) & 432 (26.0\%) & 844 (50.8\%) & 299 (18.0\%) & 85 (5.1\%) & 1,660 \\
SS posts, N & 0 & 1,128 & 1,255 & 1,455 & 3,838 \\
Total posts, med (IQR) & 14 (7--27) & 18 (9--35) & 51 (27--81) & 122 (58--235) & 22 (10--47) \\
Timeline days, med (IQR) & 636 (332--1,316) & 608 (314--1,283) & 1,006 (478--1,716) & 1,316 (833--2,309) & 726 (347--1,454) \\
\hline
\multicolumn{6}{p{13cm}}{\footnotesize\textit{Note:} SS = self-stigma. Med = median. Timeline days = days between first and last post.} \\
\end{tabularx}
\end{table*}
 
Because expression frequency was strongly associated with overall posting volume, we do not draw inferences about qualitative differences between user groups; users who wrote more posts had more opportunities to express self-stigma. Self-stigma rates varied across communities, with recovery-oriented subreddits showing the highest rates (\textit{r/OpiatesRecovery}: 8.7\%, \textit{r/Drugs}: 8.6\%) and hallucinogen and cannabis communities showing the lowest (\textit{r/LSD}: 2.6\%, \textit{r/trees}: 0.6\%). 

\subsection*{Indicator prevalence}

\begin{table*}[ht]
\centering
\caption{Codebook: Self-Stigma Indicators}
\label{tab:codebook}
\begin{tabularx}{\textwidth}{@{}p{1.5cm}p{2.5cm}p{6.5cm}p{3.5cm}p{1.0cm}@{}}
\hline
\textbf{Domain} & \textbf{Indicator} & \textbf{Definition} & \textbf{Example} & \textbf{N (\%)} \\
\hline
\multirow{9}{1.8cm}{Cognitive (Core)} 
& Self-Labeling & Self-application of stigmatized identity labels, traits, or status markers related to substance use & ``I'm just another junkie who can't get clean'' & 2,150 (56.0) \\
\cline{2-5}
& Pessimism/ Self-Defeatism & Belief that change is unlikely or impossible, ranging from general negative outlook to identity-based attribution of failure & ``I'm mentally weak. I always end up back at 4mg or more'' & 1,074 (28.0) \\
\cline{2-5}
& Deservingness/ Worthlessness & Perception of not deserving care, love, or recovery; existential disqualification from help or empathy & ``I don't deserve to be happy or have a normal life after what I've done'' & 286 (7.5) \\
\hline
\multirow{9}{1.8cm}{Affective (Core)} 
& Shame & Expressions of embarrassment, humiliation, or disgust about one's identity as a substance user & ``I'm so ashamed of what I've become. I can't even look at myself'' & 470 (12.2) \\
\cline{2-5}
& Guilt/ Self-Blame & Sadness about negative outcomes or past decisions related to substance use, with moral self-blame & ``I feel terrible for disappointing everyone who believed in me'' & 598 (15.6) \\
\cline{2-5}
& Despair/ Hopelessness & Expressions of having no value, being hopeless, or having no future tied to substance use identity & ``There's no point anymore. I'll never be anything but this'' & 1,863 (48.5) \\
\hline
\multirow{12}{1.8cm}{Behavioral} 
& Concealment & Hiding substance use from others due to fear of judgment or consequences & ``Nobody knows about my addiction. I keep it completely hidden from my family'' & 476 (12.4) \\
\cline{2-5}
& Anticipated Rejection & Expectation of negative social consequences if substance use were known & ``If my employer found out, I'd lose everything'' & 985 (25.7) \\
\cline{2-5}
& Desire to Quit & Expressed motivation to stop or reduce use, often framed as escaping a stigmatized identity & ``I just want to be normal again. I'm so tired of living like this'' & 1,642 (42.8) \\
\cline{2-5}
& Ambivalence & Conflicting feelings about substance use or recovery, often oscillating between wanting to quit and wanting to continue & ``Part of me wants to get clean but another part knows I'll just go back'' & 1,521 (39.6) \\
\hline
\multicolumn{5}{p{15.5cm}}{\footnotesize\textit{Note:} N (\%) = prevalence among 3,838 self-stigma posts. Indicators are not mutually exclusive; posts may contain multiple indicators. Examples are paraphrased composites to protect user privacy.} \\
\end{tabularx}
\end{table*}

Indicator classifiers were applied to all 72,115 posts (\S~\ref{subsec:data}). We first report indicator prevalence within the 3,838 self-stigma posts to characterize the composition of self-stigma expression (Table~\ref{tab:indicator_prevalence}); we then compare these prevalences against the 68,277 non-self-stigma posts as a test of indicator distinctiveness. Within self-stigma posts, Self-Labeling was the most prevalent indicator (56.0\% of posts), followed by Despair/Hopelessness (48.5\%), Desire to Quit (42.8\%), and Ambivalence (39.6\%). Shame (12.2\% of posts, 23.6\% of users) and Guilt/Self-Blame (15.6\% of posts, 30.0\% of users) were less prevalent at the post level but were expressed by a substantial subset of users. Deservingness/Worthlessness was least common (7.5\% of posts, 14.8\% of users), representing the most severe form of cognitive self-stigma involving beliefs of existential disqualification from care or happiness.
 
At the user level, prevalence was uniformly higher than at the post level, reflecting repeated expression among users who articulate a given indicator. Among the 1,228 users with at least one self-stigma expression, the most frequently endorsed indicators were Self-Labeling (72.1\%), Despair/Hopelessness (61.0\%), and Desire to Quit (60.7\%).

All ten indicators were significantly more prevalent in self-stigma posts than in non-self-stigma posts after Benjamini--Hochberg FDR correction (all $p_{\text{FDR}} < .001$), with relative risks (RR) ranging from 3.6 to 86.2 (Table~\ref{tab:indicator_prevalence}). The largest differentials appeared for Deservingness/Worthlessness (RR = 86.2, 95\% CI [65.3, 113.9]), Shame (RR = 41.6 [35.4, 48.9]), and Pessimism/Self-Defeatism (RR = 29.5 [26.9, 32.3]); the smallest for Desire to Quit (RR = 3.6 [3.4, 3.7]) and Concealment (RR = 6.2 [5.6, 6.9]).

\begin{table*}[ht]
\centering
\caption{Prevalence of Self-Stigma Indicators}
\label{tab:indicator_prevalence}
\begin{tabular}{llccc}
\hline
\textbf{Domain} & \textbf{Indicator} & \textbf{SS Posts \textit{N} (\%)} & \textbf{Users \textit{N} (\%)} & \textbf{RR [95\% CI]} \\
\hline
\multirow{3}{*}{Cognitive} 
    & Self-Labeling & 2,150 (56.0) & 885 (72.1) & 12.4 [11.9, 13.0] \\
    & Pessimism/Self-Defeatism & 1,074 (28.0) & 536 (43.6) & 29.5 [26.9, 32.3] \\
    & Deservingness/Worthlessness & 286 (7.5) & 182 (14.8) & 86.2 [65.3, 113.9] \\
\hline
\multirow{3}{*}{Affective} 
    & Shame & 470 (12.2) & 290 (23.6) & 41.6 [35.4, 48.9] \\
    & Guilt/Self-Blame & 598 (15.6) & 369 (30.0) & 13.4 [12.1, 14.9] \\
    & Despair/Hopelessness & 1,863 (48.5) & 749 (61.0) & 19.0 [18.0, 20.1] \\
\hline
\multirow{4}{*}{Behavioral} 
    & Concealment & 476 (12.4) & 276 (22.5) & 6.2 [5.6, 6.9] \\
    & Anticipated Rejection & 985 (25.7) & 491 (40.0) & 9.4 [8.8, 10.1] \\
    & Desire to Quit & 1,642 (42.8) & 745 (60.7) & 3.6 [3.4, 3.7] \\
    & Ambivalence & 1,521 (39.6) & 694 (56.5) & 9.2 [8.7, 9.7] \\
\hline
\multicolumn{5}{p{14cm}}{\footnotesize\textit{Note:} SS Posts (\textit{n} = 3,838) and Users (\textit{n} = 1,228) prevalence; RR = relative risk versus non-self-stigma posts (\textit{n} = 68,277), 95\% CI in brackets. All chi-square tests significant after Benjamini--Hochberg FDR correction (all $p_{\text{FDR}} < .001$, effectively zero at this sample size; interpret via RR/CI).} \\

\end{tabular}
\end{table*}

\subsection*{Core and behavioral indicator co-occurrence}
 
Self-stigma theory distinguishes between core cognitive/affective components and their behavioral impacts \citep{Corrigan2010social, corriganSelfStigmaMentalIllness2012}. We examined whether this theoretical distinction is reflected in how users articulate self-stigma in naturalistic discourse by analyzing co-occurrence patterns at the user level, the unit of analysis most consistent with theoretical models describing individual psychological processes.
 
\paragraph*{User-level association}
 
Among the 1,228 users who expressed self-stigma, 76.1\% ($n$ = 935) expressed both core and behavioral indicators at some point in their posting history, while only 5.0\% ($n$ = 61) expressed behavioral indicators without ever expressing core indicators (Table~\ref{tab:user_cooccurrence}). The association between core and behavioral expression was strong: 84.0\% of users who expressed core indicators also expressed behavioral indicators, compared to 53.0\% of users who never expressed core indicators (OR = 4.65, 95\% CI [3.12, 6.94]; Fisher's exact $p < 0.001$). This association was also evident within individual posts. Among the 2,663 self-stigma posts containing any behavioral indicator, 87.0\% ($n$ = 2,318) also contained at least one core indicator, indicating that behavioral manifestations rarely appeared in isolation from the internalized beliefs theorized to produce them.

\begin{table}[ht]
\centering
\caption{User-Level Co-occurrence of Core and Behavioral Indicators ($N$ = 1,228 Users)}
\label{tab:user_cooccurrence}
\footnotesize
\begin{tabular}{lccc}
\hline
 & \multicolumn{2}{c}{\textbf{Behavioral}} & \\
\cline{2-3}
\textbf{Core} & \textbf{Never} & \textbf{Ever} & \textbf{Total} \\
\hline
Never & 54 (4.4\%) & 61 (5.0\%) & 115 (9.4\%) \\
Ever & 178 (14.5\%) & 935 (76.1\%) & 1,113 (90.6\%) \\
\hline
Total & 232 (18.9\%) & 996 (81.1\%) & 1,228 (100\%) \\
\hline
\multicolumn{4}{p{7.5cm}}{\textit{Note:} ``Ever'' = expressed in at least one self-stigma post. OR = 4.65 [3.12, 6.94], Fisher's exact $p < .001$.} \\
\end{tabular}
\end{table}

\paragraph*{Pairwise indicator associations}
 
\begin{figure}[!ht]
    \centering
    \includegraphics[width=.75\linewidth]{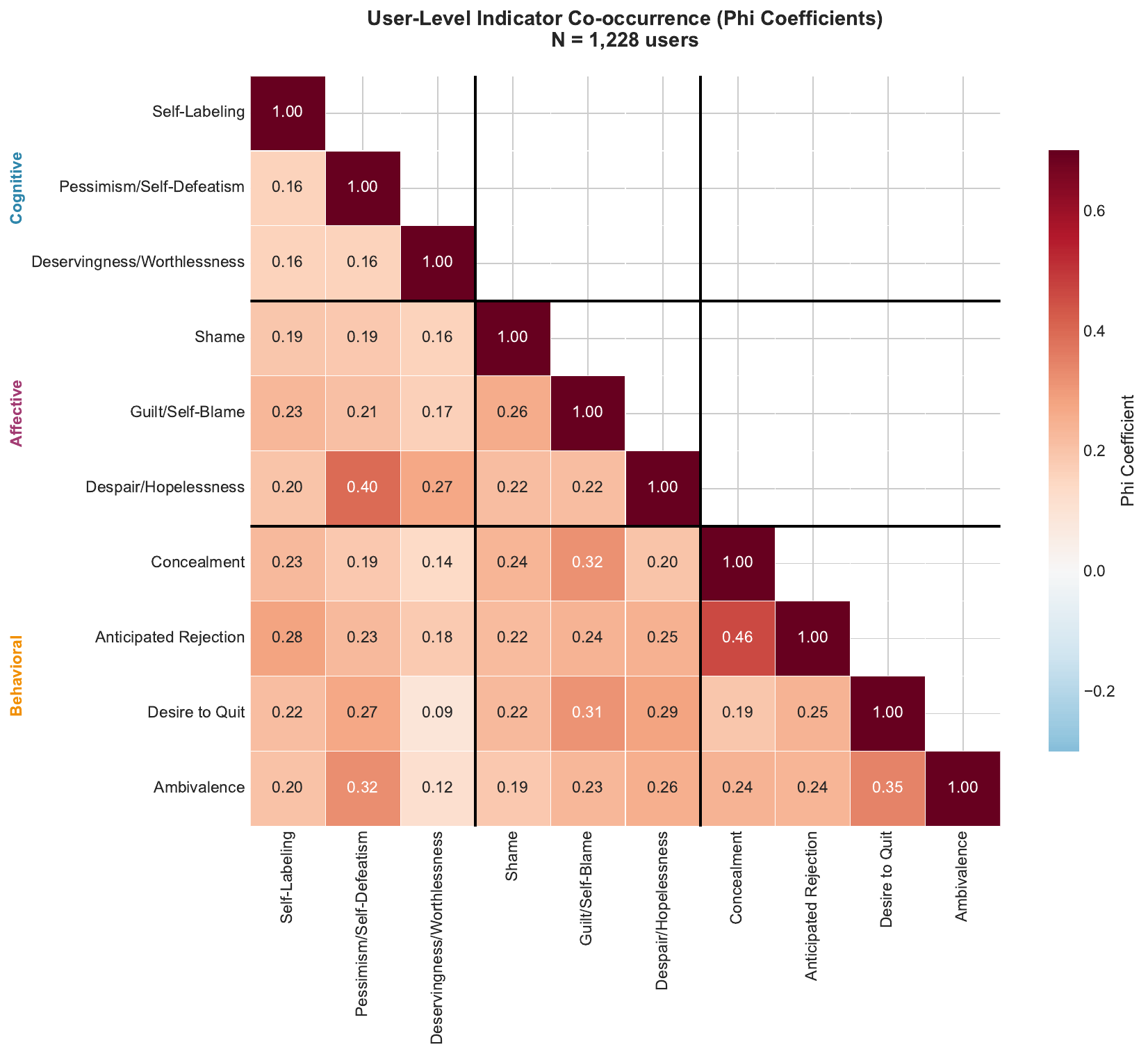}
    \caption{User-level indicator co-occurrence shown as phi coefficients ($N$ = 1,228 users). All pairwise associations were significant after FDR correction ($p < 0.01$). Black lines separate indicator domains. The strongest associations appeared within the behavioral domain (Concealment--Anticipated Rejection, $\phi$ = 0.46) and between cognitive and affective indicators of hopelessness (Pessimism--Despair, $\phi$ = 0.40).}
    \label{fig:cooccurrence_heatmap}
\end{figure}

To examine which specific indicators co-occurred, we calculated pairwise associations at the user level using phi coefficients and odds ratios (Figure~\ref{fig:cooccurrence_heatmap}). All 45 indicator pairs showed significant positive associations after FDR correction (all $p_{\text{FDR}} < 0.01$; full results in Appendix \ref{supp:pairwise_or}). The strongest associations emerged within the behavioral domain: Concealment and Anticipated Rejection showed the highest co-occurrence ($\phi$ = 0.46, OR = 12.0), and Desire to Quit and Ambivalence were also strongly linked ($\phi$ = 0.35, OR = 4.4). 
 
Cross-domain associations between core and behavioral indicators were consistently moderate to strong. Guilt/Self-Blame showed particularly robust associations with behavioral indicators, including Desire to Quit (OR = 5.0), Concealment (OR = 4.7), and Anticipated Rejection (OR = 2.9). Self-Labeling was similarly associated with Concealment (OR = 5.0) and Anticipated Rejection (OR = 4.2). Among affective-cognitive pairs, Pessimism/Self-Defeatism and Despair/Hopelessness showed strong co-occurrence ($\phi$ = 0.40, OR = 6.2), as did Deservingness/Worthlessness and Despair/Hopelessness (OR = 9.3). 
These indicators were retained as distinct in coding because they target different theoretical mechanisms---a cognitive belief about the possibility of change (Pessimism), an affective state of futurelessness (Despair), and existential disqualification from care or recovery (Deservingness).

\FloatBarrier

 
\subsection*{Temporal patterns in self-stigma expression}
 
To examine how self-stigma expression unfolds over time, we restricted the analyses to the 384 users with three or more self-stigma posts, representing 31.3\% of users who expressed self-stigma. These users contributed 2,710 self-stigma posts and maintained median posting timelines of 1,316 days (IQR: 700--2,291), with a median of 18 days between consecutive self-stigma posts. 
 
We note throughout that emergence positions reflect when users first \textit{disclosed} each indicator in text rather than when they first experienced the underlying psychological state. The temporal patterns described below should be read as patterns of online expression and community engagement, not as trajectories of psychological development.
 
\begin{figure}[!ht]
    \centering
    \includegraphics[width=.75\linewidth]{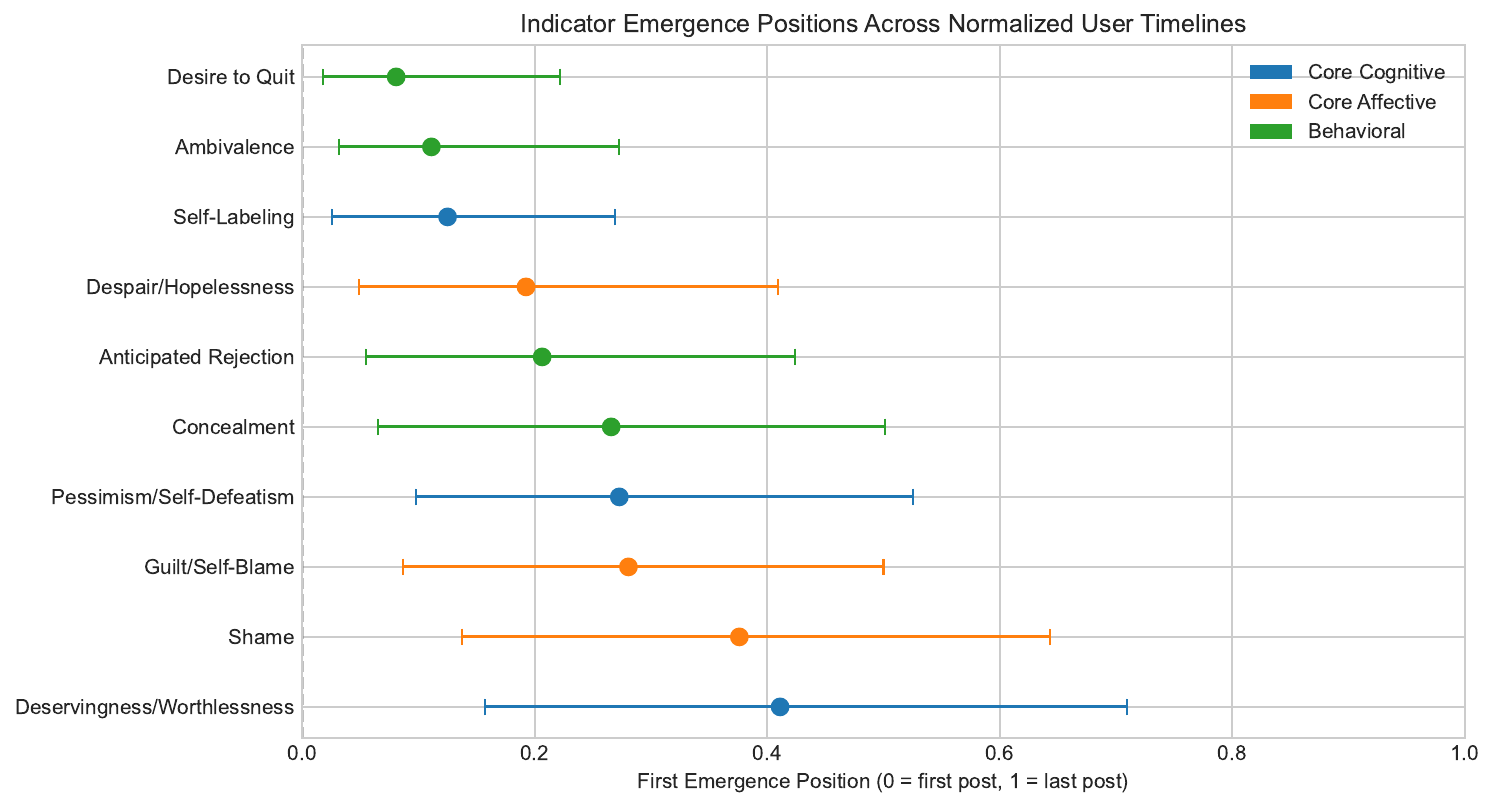}
    \caption{Median emergence positions for self-stigma indicators across users' posting histories. Indicators are ordered by median first-appearance position (0 = first post, 1 = last post). Error bars represent interquartile range. 
    }
    \label{fig:emergence}
\end{figure}

\paragraph*{Indicator emergence}
We first examined when each indicator first appeared in users' posting histories (Table~\ref{tab:emergence}, Figure \ref{fig:emergence}). Behavioral indicators emerged earliest (Desire to Quit at 0.081; Ambivalence at 0.111), followed closely by Self-Labeling (0.125), the earliest-emerging cognitive indicator. Core affective indicators emerged later, with Despair/Hopelessness at 0.192 and Shame at 0.376. The least prevalent indicators, Guilt/Self-Blame (0.281) and Deservingness/Worthlessness (0.411), emerged latest. 

\begin{table}[ht]
\centering
\caption{Emergence Position of Self-Stigma Indicators}
\label{tab:emergence}
\footnotesize
\begin{tabular}{l p{2.5cm} cc}
\hline
\textbf{Domain} & \textbf{Indicator} & \textbf{N (\%)} & \textbf{Median [IQR]} \\
\hline
Behavioral & Desire to Quit & 372 (96.9) & 0.081 [0.02, 0.22] \\
Behavioral & Ambivalence & 359 (93.5) & 0.111 [0.03, 0.27] \\
Cognitive & Self-Labeling & 375 (97.7) & 0.125 [0.03, 0.27] \\
Affective & Despair/Hopelessness & 360 (93.8) & 0.192 [0.05, 0.41] \\
Behavioral & Anticipated \newline Rejection & 336 (87.5) & 0.206 [0.06, 0.42] \\
Behavioral & Concealment & 270 (70.3) & 0.266 [0.07, 0.50] \\
Cognitive & Pessimism/Self-Defeatism & 311 (81.0) & 0.273 [0.10, 0.53] \\
Affective & Guilt/Self-Blame & 265 (69.0) & 0.281 [0.09, 0.50] \\
Affective & Shame & 188 (49.0) & 0.376 [0.14, 0.64] \\
Cognitive & Deservingness/\newline Worthlessness & 114 (29.7) & 0.411 [0.16, 0.71] \\
\hline
\multicolumn{4}{p{8cm}}{\textit{Note:} Position normalized within each user's posting sequence (0 = first post, 1 = last post). N (\%) = users who ever expressed this indicator. Ordered by median emergence position.} \\
\end{tabular}
\end{table}

\paragraph*{Core versus behavioral sequencing} 
Among the 381 users who expressed at least one indicator from each domain (three users expressed indicators from only one domain), first appearance fell into three patterns: 37.5\% ($n$ = 143) expressed core and behavioral indicators in the same post (concurrent disclosure), 38.6\% ($n$ = 147) first expressed a behavioral indicator in an earlier post than any core indicator (behavioral-first), and 23.9\% ($n$ = 91) first expressed a core indicator in an earlier post than any behavioral indicator (core-first). 
Among the 238 users whose indicators emerged in separate posts, the behavioral-first pattern occurred roughly 1.6 times as often as the core-first pattern (binomial test against equal probability, $p$ < 0.001). This asymmetry was consistent across expression frequency groups.

\FloatBarrier

\paragraph*{Stability of self-stigma expression over time}
 
Having established when indicators first appeared, we examined whether the breadth of self-stigma expression changed across users' posting trajectories. We computed the number of co-occurring indicators per self-stigma post and fitted a GEE model testing whether this count varied with normalized posting position. The number of indicators per post remained stable across trajectories (mean = 2.92, SD = 1.76; GEE $\beta$ = 0.067, SE = 0.117, $p$ = 0.57), with LOESS-smoothed estimates showing essentially no change from first post (2.75, 95\% CI [2.55, 2.94]) to last post (2.78, 95\% CI [2.62, 2.98]). This pattern held in both the Moderate (3--7 self-stigma posts; $\beta$ = 0.136, $p$ = 0.32) and High ($\geq$8 posts; $\beta$ = $-$0.020, $p$ = 0.92) expression frequency subgroups.

\begin{figure}[!ht]
    \centering
    \includegraphics[width=.75\linewidth]{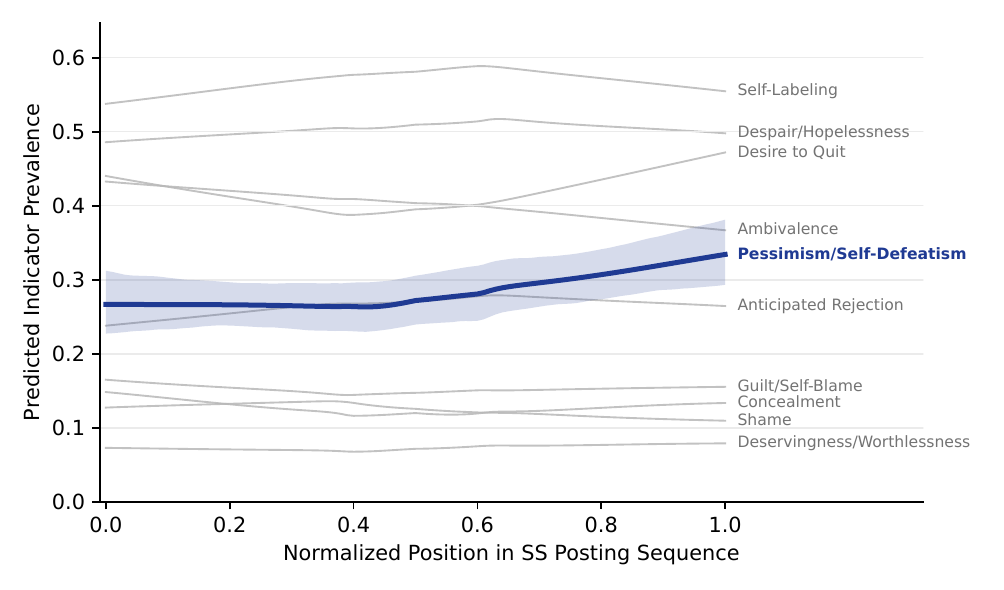}
    \caption{LOESS-smoothed prevalence of all ten self-stigma indicators across users' normalized posting trajectories ($N$ = 2,710 posts from 384 users). Pessimism/Self-Defeatism (bold, with shaded 95\% CI band from cluster bootstrap; 500 user-level resamples) was the only indicator showing significant temporal change (OR = 1.62, 95\% CI [1.25, 2.10], $p_{\text{FDR}}$ = 0.002). 
    }
    \label{fig:pessimism_trajectory}
\end{figure}
\FloatBarrier

Consistent with this overall stability, individual indicator GEE models showed no significant temporal change for nine of ten indicators after FDR correction (Appendix Table \ref{tab:supp_gee_temporal}). The sole exception was Pessimism/Self-Defeatism, which showed modestly higher odds of expression in later versus earlier posts (OR = 1.62, 95\% CI [1.25, 2.10], $p_{\text{FDR}} = 0.002$; Figure~\ref{fig:pessimism_trajectory}). This finding was consistent across expression frequency subgroups (Moderate: OR = 1.29; High: OR = 1.28). Comparisons of indicator prevalence in users' first versus last self-stigma posts similarly showed no significant changes after correction (all $p_{\text{FDR}} > 0.19$; see Appendix Table \ref{tab:supp_first_last}). A complementary analysis examining the compositional structure of self-stigma expression across consecutive posts is reported in Appendix \ref{supp:robustness}.
\section*{Discussion}\label{discussion}

This study developed a framework to identify and measure self-stigma expression among PWUD in online substance use communities, and applied it to 72,115 Reddit posts from 1,660 users to examine whether and how these expressions shift across user timelines. Three patterns were prevalent: self-stigma expression was integrated rather than componential, with behavioral indicators rarely appearing without core cognitive or affective ones; behavioral indicators tended to emerge before or alongside core ones in users' timelines; and expression patterns remained largely stable over time, with the exception of Pessimism/Self-Defeatism, which increased. We interpret these patterns as features of disclosure in online substance use communities rather than direct evidence about underlying psychological processes.

The codebook captured the cognitive, affective, and behavioral content that self-stigma theory describes \citep{corriganSelfStigmaMentalIllness2012, luomaSelfStigmaSubstanceAbuse2013, earnshawStigmaSubstanceUse2020}. The way these indicators co-occurred also mapped onto theorized pathways: the strong link between Concealment and Anticipated Rejection corroborates Modified Labeling Theory's account of secrecy as a protective response to anticipated devaluation \citep{linkModifiedLabelingTheory1989}, and Self-Labeling co-occurred robustly with both Concealment and Anticipated Rejection. The strong user-level associations and the rarity of behavioral indicators occurring without core indicators provide empirical support for the theoretical distinction between these domains. However, co-occurrence is also where our findings begin to push against an assumption embedded in much of the existing measurement literature: that behavioral manifestations of self-stigma are downstream consequences of internalized cognitive and affective beliefs \cite{corrigan2006self,corrigan2009whytry, corriganSelfStigmaMentalIllness2012}. Progressive models position behavior as the endpoint of a sequence that begins with cognitive endorsement and affective response and concludes in concealment, withdrawal, and disengagement from valued goals \citep{corrigan2006self, corriganSelfStigmaMentalIllness2012}. In naturalistic text, this ordering is difficult to recover. Posts often interleave behavioral and core content in a single utterance; a statement such as \textit{``Part of me wants to get clean but another part knows I'll just go back''} embeds motivational behavior, pessimistic cognition, and the affective weight of repeated failure in one instance. Whether the behavior precipitates the cognition, follows from it, or sustains it cannot be parsed from the text. In disclosures of this kind, whether on social media or in other written records such as diaries or journals, self-stigma may be better understood as cyclical than linear, with concealment under fear of judgment reinforcing the self-label, which sustains the affective burden, which in turn sustains the behavior. The patterns observed here suggest that the linear framing common in survey-based stigma measurement \citep{luomaSelfStigmaSubstanceAbuse2013, smithSubstanceUseStigma2016} may understate the recursive dynamics of expressed self-stigma.
 
Two temporal patterns warrant closer interpretation. First, indicators representing the most severe forms of internalized stigma—such as Deservingness/Worthlessness, which involves beliefs of existential disqualification from care, and Guilt/Self-Blame—were among the least prevalent and the slowest to surface in users' timelines (median emergence positions of 0.41 and 0.28, respectively). Shame is theoretically central to self-stigma \citep{luomaSelfStigmaSubstanceAbuse2013, smithSubstanceUseStigma2016}, yet it was similarly slow to surface. We do not read this as evidence that affective or severe cognitive self-stigma develops late. We read it as evidence that it is the hardest to articulate when writing. Substance use subreddits function as something like asynchronous, pseudonymous peer-support spaces \citep{macleanForum77AnalysisOnline2015, almeida_medication_2025, krawczyk_self_2025}, and in any such setting, the entry-level register tends to be motivational and informational rather than identity-level affective. Naming oneself as ashamed requires accumulated trust; naming oneself as wanting to quit does not \citep{dagostinoSocialNetworkingOnline2017a,yangSelfdisclosureChannelDifference2017}. That a substantial subset of users eventually did articulate shame and guilt (23.6\% and 30.0\% at the user level) suggests these spaces afford disclosure of the affective core over time, even if not on entry. This pattern echoes the strategic disclosure that contact-based stigma-reduction programs aim to facilitate \citep{corrigan2025decreasing, earnshawDisclosureProcessesPredictors2021}: users may titrate what to share and when, and the affective dimensions of self-stigma appear to surface later in users' posting sequences, consistent with the possibility that disclosure becomes more tenable as posting history accumulates.
 
The one indicator that did shift systematically over time was Pessimism/Self-Defeatism, which became more prevalent in users' later posts. This is the indicator most closely aligned with Corrigan's ``why try'' effect, the diminished sense that recovery is worth attempting \citep{corrigan2009whytry, corrigan2016impact}, and statements such as \textit{``I don't think I can attempt recovery again''} became more characteristic later in users' trajectories. We note that ``change'' here is not limited to cessation; pessimism may reflect diminished expectations about a range of valued life goals, including harm reduction, relationship repair, or improved well-being, consistent with the original framing of the ``why try'' effect as encompassing life goals broadly rather than abstinence alone \citep{corrigan2009whytry}. We do not interpret this as evidence that self-stigma itself intensifies; the other nine indicators were stable, though descriptive trends added texture without reaching statistical significance: Desire to Quit showed a directional increase, while Guilt/Self-Blame and Ambivalence both trended downward. Larger samples would be needed to determine whether these patterns reflect genuine compositional shifts alongside the deepening of pessimism. What appears to intensify is a specific belief about the possibility of change, one that may be amplified by recurrent experience within these communities, including witnessing peers relapse, returning to use oneself, or watching others taper and fail. The selective deepening of pessimism alongside stability in other indicators is consistent with theoretical accounts that position the ``why try'' effect as a distinct mechanism by which self-stigma forecloses engagement with recovery \citep{corrigan2009whytry, corriganSelfStigmaMentalIllness2012}. One possibility worth flagging is that pessimism may be most acute in communities where recovery is implicitly framed as a desired outcome (e.g., \textit{r/OpiatesRecovery}, \textit{r/StopSpeeding}) rather than as an ongoing process; the abstinence-violation effect \citep{marlatt2005relapse} of each return to use may deepen pessimistic beliefs in ways a harm-reduction framing might not \citep{corrigan2025decreasing}.
 
These findings suggest several directions for addiction research and practice. First, the early emergence of motivational expression (Desire to Quit, Ambivalence) suggests that online substance use communities may capture PWUD at a point of nascent change motivation, consistent with prior work documenting online health communities as entry points for individuals contemplating behavior change \citep{macleanForum77AnalysisOnline2015}. These communities reach PWUD whose engagement with formal care is unknown or absent \citep{SAMHSA2024} and represent a touchpoint that stigma-reduction efforts have not systematically used. Second, the selective increase in Pessimism/Self-Defeatism over time indicates that interventions targeting beliefs about the possibility of change may be most effective when delivered before such beliefs become entrenched. Third, this may carry particular weight given the elevated risk of self-harm among people with SUD, for which prior self-harm, psychiatric comorbidity, and despair-related cognitions are established predictors \citep{yu_self_2026}. The indicators that persist across users' trajectories (Despair/Hopelessness, Deservingness/Worthlessness) or intensify (Pessimism/Self-Defeatism) in our sample overlap conceptually with this risk profile; online expression of these indicators may warrant attention as a complementary signal alongside structured clinical assessment.

\subsection*{Strengths, limitations, and future directions}
 
This study has several strengths. It is, to our knowledge, the first to characterize self-stigma expression longitudinally in naturalistic discourse from PWUD, at a scale (72,115 posts, 1,660 users, 19 years) and across a population broader than typical treatment-seeking samples. The hybrid coding approach, anchored in expert human consensus and validated against held-out data, demonstrates that theory-informed indicators of self-stigma can be reliably identified in unstructured text.
 
Several limitations bound interpretation. First, this is a discourse study. The temporal patterns we observe reflect when users disclosed particular experiences, which need not correspond to when those experiences emerged psychologically. Second, post-position rather than calendar time anchors the temporal analyses; this was necessary given variable posting cadence but precludes claims about real-time pace. Third, our sample over-represents users with sustained engagement and opioid-related communities (35.9\% of posts), and lacks demographic information that would allow examination of how self-stigma expression varies across PWUD subgroups \citep{hatzenbuehlerStigmaFundamentalCause2013}. Fourth, the classifier's high recall (1.000) and moderate precision (0.670) mean estimates likely overstate true prevalence; we report patterns within identified expressions rather than population-based rates. 

Future work should test whether the recursive dynamics suggested here can be confirmed by pairing online discourse with sequential self-report or ecological momentary assessment from the same individuals \citep{earnshawIntegratingTimeStigma2022, fox2025advancing}, apply the indicator framework to other stigmatized conditions, and examine whether the temporal patterns observed here generalize to other platforms and communities.

\section*{Conclusion}\label{conclusion}

Self-stigma is expressed by PWUD in online substance use communities as an integrated cognitive, affective, and behavioral phenomenon, with patterns of disclosure shaped by the communicative norms of these communities rather than by a fixed developmental sequence. Behavioral indicators tend to surface first, expression patterns remain largely stable once established, and pessimism about the possibility of change deepens over time. Naturalistic discourse from these communities offers a complementary window onto self-stigma as it is voiced by PWUD on their own terms, and points to online spaces as a potential site for early stigma-reduction efforts.

\subsection*{Declaration of interest}
None.

\subsection*{Data availability statement}
The analysis code and LLM prompts used in this study will be made available on GitHub upon acceptance. The Reddit post data cannot be shared publicly due to user privacy considerations, consistent with the pseudonymous nature of the platform and ethical guidelines for social media research. Derived aggregate data supporting the findings may be made available from the corresponding author upon reasonable request.

\bibliographystyle{cas-model2-names}
\bibliography{references,more_references}

\appendix


\section{Model Selection and Validation}\label{supp:validation_metrics}

\subsection{Model Comparison on Development Set}\label{supp:model_comparison}

We evaluated four language models on the 350-post development set to select the optimal classifier for self-stigma detection. All models used identical prompts with three-run majority voting to reduce classification variance. Table~\ref{tab:supp_model_comparison} presents performance metrics for each model.

\begin{table*}[ht!]
\centering
\caption{Comparison of Language Model Performance on Self-Stigma Classification}
\label{tab:supp_model_comparison}
\resizebox{\textwidth}{!}{%
\footnotesize
\begin{tabular}{lcccccccc}
\hline
\textbf{Model} & \textbf{Raw Agreement} & \textbf{Kappa} & \textbf{F1} & 
\textbf{Precision} & \textbf{Recall} & \textbf{Accuracy} & \textbf{Cost In/Out per 1M}$^a$ \\
\hline
GPT-4.1-nano & 0.791 & 0.533 & 0.676 & 0.874 & 0.551 & 0.791 & \$0.05 / \$0.20 \\
GPT-4.1-mini & 0.891 & 0.769 & 0.856 & 0.897 & 0.819 & 0.891 & \$0.20 / \$0.80 \\
GPT-4o-mini & 0.866 & 0.716 & 0.824 & 0.853 & 0.797 & 0.866 & \$0.075 / \$0.30 \\
Llama-3.3-70B-Instruct-Turbo$^b$ & 0.860 & 0.714 & 0.835 & 0.780 & 0.899 & 0.860 & \$0.88 / \$0.88 \\
Qwen2.5-7B-Instruct-Turbo$^b$ & 0.794 & 0.593 & 0.776 & 0.679 & 0.906 & 0.794 & \$0.30 / \$0.30 \\
DeepSeek-V3.1$^b$ & 0.871 & 0.726 & 0.829 & 0.872 & 0.790 & 0.871 & \$0.60 / \$1.70 \\
Gemini 2.5 Flash$^g$ & 0.797 & 0.584 & 0.758 & 0.716 & 0.804 & 0.797 & \$0.30 / \$2.50 \\
\hline
\multicolumn{8}{p{\linewidth}}{ \textit{Note:} Development set consisted of 350 posts 
(150 positive, 200 negative for self-stigma) coded by two human raters with consensus 
resolution. All models used three-run majority voting with temperature=0.0.} \\
\multicolumn{8}{p{\linewidth}}{\small $^a$Cost estimates based on OpenAI API pricing as of 
12/08/2025 for prompt + completion tokens. Does not include batch API discounts.} \\
\multicolumn{8}{p{\linewidth}}{\small $^b$Inference performed on Together.ai.} \\
\multicolumn{8}{p{\linewidth}}{\small $^g$Inference performed on Google Gemini API} \\
\end{tabular}
}
\end{table*}

GPT-4.1-mini was selected based on its superior balance of inter-rater agreement (substantial kappa of 0.769), F1 score (0.856), and cost-effectiveness. Among the alternatives, GPT-4o-mini achieved comparable accuracy at lower per-token cost but showed slightly lower agreement ($\kappa = 0.716$) and F1 (0.824), differences that compounded across 72,115 posts classified in triplicate. The open-weight models Llama-3.3-70B and DeepSeek-V3.1 achieved similar kappa values (0.714 and 0.726, respectively) but exhibited different error profiles: Llama-3.3-70B favored recall (0.899) over precision (0.780), while DeepSeek-V3.1 favored precision (0.872) over recall (0.790). GPT-4.1-nano and Qwen2.5-7B, the two smallest models evaluated, showed the weakest agreement ($\kappa = 0.533$ and $0.593$), suggesting that self-stigma classification requires sufficient model capacity to distinguish identity-level self-devaluation from related but distinct constructs such as situational distress and behavioral regret. 

\subsection{Post-Hoc Validation Results}

To verify classifier performance on truly independent data not used during prompt development, we conducted post-hoc validation on a stratified random sample of 300 posts (100 GPT-positive, 200 GPT-negative). The classifier maintained substantial agreement with expert human coding (Cohen's $\kappa = 0.730$, 95\% CI [.727, .865]), with validation metrics presented in Table~\ref{tab:supp_validation_metrics}.

\begin{table}[ht!]
\centering
\caption{Self-Stigma Classification Performance on Held-Out Validation Set (N=300)}
\label{tab:supp_validation_metrics}
\begin{tabular}{lc}
\hline
\textbf{Metric} & \textbf{Value} \\
\hline
Cohen's Kappa & 0.730 \\
F1 Score & 0.802 (95\% CI: [0.727, 0.865]) \\
Precision & 0.670 \\
Recall & 1.000 \\
Matthews Correlation Coefficient & 0.758 \\
Raw Agreement & 89.0\% \\
\hline
\multicolumn{2}{l}{\textbf{Confusion Matrix:}} \\
\quad True Positives & 67 \\
\quad False Positives & 33 \\
\quad False Negatives & 0 \\
\quad True Negatives & 200 \\
\hline
\multicolumn{2}{l}{\textit{Note:} Validation sample stratified with 100 GPT-positive,} \\
\multicolumn{2}{l}{200 GPT-negative posts. Bootstrap CIs based on 1,000 iterations.} \\
\end{tabular}
\end{table}

The classifier demonstrated perfect recall (1.000), capturing all true self-stigma expressions in the validation sample, but moderate precision (0.670). McNemar's test revealed significant systematic bias toward over-classification ($p < .001$), with GPT-4.1-mini producing 33 false positives but zero false negatives. This asymmetric error pattern indicates the model is highly sensitive to potential self-stigma markers but occasionally misclassifies related constructs. Given the classifier's precision characteristics, we estimate true prevalence at approximately 3.5\% of posts.

\section{Indicator-Level Classification Validation}\label{supp:indicator_validation}

\subsection{Indicator Codebook Validation}\label{supp:irr_results}

Two coders independently applied the 10-indicator codebook to a stratified sample of 50 posts from substance use-related subreddits. Inter-rater reliability was assessed using Cohen's kappa for each indicator. Seven indicators achieved at least substantial agreement ($\kappa \geq$ 0.6), demonstrating strong codebook clarity. Three indicators with initial agreement below the recommended threshold ($\kappa < $ 0.60): Pessimism/Self-Defeatism, Deservingness/Worthlessness, and Ambivalence, underwent independent adjudication by a third coder blind to the original coders' decisions. Following adjudication, all three indicators achieved substantial agreement ($\kappa$ = 0.500--0.787). The overall macro-averaged kappa of 0.723 indicates substantial reliability across all indicators. For non-adjudicated indicators, disagreements were resolved through consensus discussion, with final labels assigned if either coder identified the indicator as present. These consensus labels served as ground truth for subsequent automated classification validation.

\begin{table}[ht]
\centering
\caption{Inter-Rater Reliability for Self-Stigma Indicators (N=50 posts)}
\label{tab:supp_irr_results}
{\small                          
\setlength{\tabcolsep}{4pt}      
\begin{tabular}{lccc}
\hline
\textbf{Indicator} & \textbf{Kappa} & \textbf{Agree.\%} & \textbf{Prev.} \\
\hline
\multicolumn{4}{l}{\textit{Core Indicators (Cognitive-Affective)}} \\
Self-Labeling                    & 0.799 & 90.0\% & 45.0\% \\
Pessimism/Self-Defeatism*        & 0.787 & 90.0\% & 36.0\% \\
Deservingness/Worthlessness*     & 0.713 & 96.0\% &  8.0\% \\
Shame                            & 0.627 & 86.0\% & 25.0\% \\
Guilt/Self-Blame                 & 0.834 & 92.0\% & 40.0\% \\
Despair/Hopelessness             & 0.642 & 82.0\% & 47.0\% \\
\cline{2-3}
\multicolumn{1}{r}{\textit{Core macro-avg:}} & 0.734 & 89.3\% & \\
\hline
\multicolumn{4}{l}{\textit{Behavioral Consequence Indicators}} \\
Anticipated Rejection            & 0.732 & 92.0\% & 18.0\% \\
Concealment                      & 0.898 & 98.0\% & 11.0\% \\
Desire to Quit                   & 0.702 & 88.0\% & 28.0\% \\
Ambivalence*                     & 0.500 & 96.0\% &  8.0\% \\
\hline
\textbf{Overall macro-averaged}  & 0.723 & 91.0\% & --     \\
\hline
\multicolumn{4}{p{\linewidth}}{\scriptsize \textit{Note:} Two independent coders.
For most indicators, kappa represents inter-coder agreement; disagreements resolved
through consensus discussion.} \\
\multicolumn{4}{p{\linewidth}}{\scriptsize *Indicators with initial $\kappa < 0.60$
were adjudicated by an expert third coder. Kappa represents average agreement between
each coder and the adjudicated gold standard.} \\
\end{tabular}
}
\end{table}

\subsection{Final Validation Results}

Table~\ref{tab:supp_indicator_performance} presents final validation metrics for each 
indicator after iterative prompt refinement.

\begin{table*}[ht!]
\centering
\caption{Per-Indicator classification performance on LLM validation set (N = 100). Samples for each indicator vary based on number of positive cases in validation set. Negative samples taken dynamically based on 2:1 ratio.}
\label{tab:supp_indicator_performance}
\small
\begin{tabular}{@{}lccccc@{}}
\hline

\textbf{Indicator} & \textbf{Kappa} & \textbf{Prevalence}$^a$ & \textbf{Precision} & \textbf{Recall} & \textbf{F1} \\
\hline
Anticipated Rejection & 0.639 & 0.283 & 0.619 & 1.000 & 0.765 \\
Concealment & 0.780 & 0.367 & 0.900 & 0.818 & 0.857 \\
Desire to Quit & 0.653 & 0.460 & 0.765 & 0.897 & 0.825 \\
Self-Labeling & 0.589 & 0.488 & 0.702 & 1.000 & 0.825 \\
Guilt/Self-Blame & 0.689 & 0.441 & 0.840 & 0.808 & 0.824 \\
Despair/Hopelessness & 0.836 & 0.613 & 0.891 & 1.000 & 0.942 \\
Pessimism/Self-Defeatism & 0.675 & 0.615 & 0.875 & 0.875 & 0.875 \\
Deservingness/Worthlessness & 0.769 & 0.238 & 0.714 & 1.000 & 0.833 \\
Ambivalence & 0.898 & 0.651 & 1.000 & 0.927 & 0.962 \\
Shame & 0.903 & 0.292 & 0.875 & 1.000 & 0.933 \\
\hline
\end{tabular}
\vspace{2pt}
\parbox{\textwidth}{\footnotesize}
\end{table*}

All ten indicators met the $F1 \geq 0.70$ quality threshold, with macro-averaged $F1 = 0.864$ and macro-averaged $\kappa = 0.744$. Performance was strongest for Ambivalence ($F1 = 0.962$, $\kappa = 0.898$) and Despair/Hopelessness ($F1 = 0.942$, $\kappa = 0.836$), both of which benefit from relatively unambiguous linguistic markers (e.g., conflicting motivational statements, expressions of futility). Shame ($F1 = 0.933$, $\kappa = 0.903$) and Pessimism/Self-Defeatism ($F1 = 0.875$, $\kappa = 0.675$) also performed well, particularly after additional backward evaluation refined decision boundaries. The lowest-performing indicator was Anticipated Rejection ($F1 = 0.765$, $\kappa = 0.639$), where the classifier occasionally conflated general social anxiety with stigma-specific expectations of rejection tied to substance use identity. Across indicators, recall tended to be higher than precision (median recall = 0.948 vs. median precision = 0.820), indicating that the classifiers, like the post-level classifier, were more likely to over-identify than to miss genuine instances. Four indicators, Pessimism/Self-Defeatism, Deservingness/Worthlessness, Ambivalence, and Shame, underwent additional backward evaluation in which misclassified cases from initial rounds were used to further refine prompt decision rules before final validation.

\section{Indicator Pairwise Associations by Pair Type}\label{supp:pairwise_or}

Table~\ref{tab:supp_pairwise_or} presents user-level associations between all pairs of self-stigma indicators. For each pair, we report the phi coefficient (a correlation measure for binary variables) and the odds ratio with 95\% confidence interval, indicating how much more likely users who expressed one indicator were to also express the other indicator at some point in their posting history.

Associations were uniformly positive and statistically significant after FDR correction, indicating pervasive co-occurrence across the self-stigma construct. The strongest associations appeared within the behavioral domain, particularly between Concealment and Anticipated Rejection ($\phi$ = 0.46, OR = 11.96), consistent with theoretical models linking secrecy behaviors to fears of social rejection. Within core indicators, the Pessimism/Self-Defeatism--Despair/Hopelessness pair ($\phi$ = 0.40, OR = 6.21) and Deservingness/Worthlessness--Despair/Hopelessness pair (OR = 9.30) showed the strongest associations, reflecting conceptual overlap in hopelessness-related constructs.

Cross-domain associations between core and behavioral indicators were consistently moderate (median OR = 2.94), empirically supporting the theoretical positioning of behavioral manifestations as linked to internalized cognitive and affective beliefs rather than occurring independently.

\begin{table*}[ht!]
\centering
\caption{Pairwise Indicator Associations at User Level ($N$ = 1,228 Users)}
\label{tab:supp_pairwise_or}
\footnotesize
\begin{tabular}{llccc}
\hline
\textbf{Indicator 1} & \textbf{Indicator 2} & \textbf{$\phi$} & \textbf{OR} & \textbf{95\% CI} \\
\hline
\multicolumn{5}{l}{\textit{Within Cognitive}} \\
Self-Labeling & Pessimism/Self-Defeatism & 0.16 & 2.15 & [1.65, 2.80] \\
Self-Labeling & Deservingness/Worthlessness & 0.16 & 3.54 & [2.20, 5.71] \\
Pessimism/Self-Defeatism & Deservingness/Worthlessness & 0.16 & 2.47 & [1.79, 3.42] \\
\hline
\multicolumn{5}{l}{\textit{Within Affective}} \\
Shame & Guilt/Self-Blame & 0.26 & 3.44 & [2.62, 4.53] \\
Shame & Despair/Hopelessness & 0.22 & 3.24 & [2.36, 4.43] \\
Guilt/Self-Blame & Despair/Hopelessness & 0.22 & 2.88 & [2.18, 3.80] \\
\hline
\multicolumn{5}{l}{\textit{Within Behavioral}} \\
Concealment & Anticipated Rejection & 0.46 & 11.96 & [8.53, 16.78] \\
Concealment & Desire to Quit & 0.19 & 2.86 & [2.09, 3.91] \\
Concealment & Ambivalence & 0.24 & 3.74 & [2.73, 5.12] \\
Anticipated Rejection & Desire to Quit & 0.25 & 2.94 & [2.29, 3.78] \\
Anticipated Rejection & Ambivalence & 0.24 & 2.74 & [2.15, 3.50] \\
Desire to Quit & Ambivalence & 0.35 & 4.35 & [3.41, 5.56] \\
\hline
\multicolumn{5}{l}{\textit{Cognitive--Affective}} \\
Self-Labeling & Shame & 0.19 & 3.40 & [2.34, 4.93] \\
Self-Labeling & Guilt/Self-Blame & 0.23 & 3.91 & [2.77, 5.51] \\
Self-Labeling & Despair/Hopelessness & 0.20 & 2.43 & [1.89, 3.13] \\
Pessimism/Self-Defeatism & Shame & 0.19 & 2.47 & [1.89, 3.24] \\
Pessimism/Self-Defeatism & Guilt/Self-Blame & 0.21 & 2.53 & [1.97, 3.25] \\
Pessimism/Self-Defeatism & Despair/Hopelessness & 0.40 & 6.21 & [4.74, 8.14] \\
Deservingness/Worthlessness & Shame & 0.16 & 2.56 & [1.84, 3.56] \\
Deservingness/Worthlessness & Guilt/Self-Blame & 0.17 & 2.55 & [1.85, 3.51] \\
Deservingness/Worthlessness & Despair/Hopelessness & 0.27 & 9.30 & [5.37, 16.13] \\
\hline
\multicolumn{5}{l}{\textit{Cognitive--Behavioral}} \\
Self-Labeling & Concealment & 0.23 & 4.95 & [3.22, 7.60] \\
Self-Labeling & Anticipated Rejection & 0.28 & 4.24 & [3.13, 5.74] \\
Self-Labeling & Desire to Quit & 0.22 & 2.64 & [2.04, 3.40] \\
Self-Labeling & Ambivalence & 0.20 & 2.51 & [1.94, 3.23] \\
Pessimism/Self-Defeatism & Concealment & 0.19 & 2.51 & [1.91, 3.31] \\
Pessimism/Self-Defeatism & Anticipated Rejection & 0.23 & 2.60 & [2.06, 3.29] \\
Pessimism/Self-Defeatism & Desire to Quit & 0.27 & 3.24 & [2.53, 4.15] \\
Pessimism/Self-Defeatism & Ambivalence & 0.32 & 4.03 & [3.15, 5.16] \\
Deservingness/Worthlessness & Concealment & 0.14 & 2.27 & [1.62, 3.18] \\
Deservingness/Worthlessness & Anticipated Rejection & 0.18 & 2.81 & [2.03, 3.88] \\
Deservingness/Worthlessness & Desire to Quit & 0.09 & 1.69 & [1.20, 2.38] \\
Deservingness/Worthlessness & Ambivalence & 0.12 & 2.00 & [1.42, 2.81] \\
\hline
\multicolumn{5}{l}{\textit{Affective--Behavioral}} \\
Shame & Concealment & 0.24 & 3.31 & [2.48, 4.42] \\
Shame & Anticipated Rejection & 0.22 & 2.88 & [2.20, 3.77] \\
Shame & Desire to Quit & 0.22 & 3.38 & [2.46, 4.63] \\
Shame & Ambivalence & 0.19 & 2.56 & [1.92, 3.42] \\
Guilt/Self-Blame & Concealment & 0.32 & 4.69 & [3.53, 6.22] \\
Guilt/Self-Blame & Anticipated Rejection & 0.24 & 2.90 & [2.25, 3.72] \\
Guilt/Self-Blame & Desire to Quit & 0.31 & 4.96 & [3.65, 6.74] \\
Guilt/Self-Blame & Ambivalence & 0.23 & 3.00 & [2.30, 3.93] \\
Despair/Hopelessness & Concealment & 0.20 & 2.96 & [2.16, 4.06] \\
Despair/Hopelessness & Anticipated Rejection & 0.25 & 3.03 & [2.35, 3.89] \\
Despair/Hopelessness & Desire to Quit & 0.29 & 3.34 & [2.63, 4.25] \\
Despair/Hopelessness & Ambivalence & 0.26 & 2.94 & [2.32, 3.73] \\
\hline
\multicolumn{5}{p{11cm}}{\footnotesize\textit{Note:} $\phi$ = phi coefficient. OR = odds ratio. All associations significant at $p < 0.001$ after FDR correction. Associations represent whether users who ever expressed Indicator 1 also ever expressed Indicator 2.} \\
\end{tabular}
\end{table*}

\section{Additional Temporal Sensitivity Analyses} \label{supp:temp_sensitivity}

\subsection{First Versus Last Comparisons}

As a complementary approach to the continuous GEE models, we compared indicator prevalence in each user's first self-stigma post to their last self-stigma post using McNemar's test for paired binary data, with Benjamini--Hochberg correction applied across all ten comparisons. This paired comparison provides a discrete test of whether users' self-stigma expression changed between the beginning and end of their observed posting histories, without assumptions about the functional form of change over time.

Table~\ref{tab:supp_first_last} presents these results. No indicator showed a statistically significant change after FDR correction (all $p_{\text{FDR}} > 0.18$). Several indicators showed directional decreases from first to last post, including Ambivalence ($-4.4$ percentage points, $p_{\text{FDR}} = 0.200$), Self-Labeling ($-4.2$ pp, $p_{\text{FDR}} = 0.210$), and Guilt/Self-Blame ($-3.4$ pp, $p_{\text{FDR}} = 0.188$), though none survived multiple comparison correction. Desire to Quit was the only indicator showing a directional increase ($+3.4$ pp), though this too was non-significant ($p_{\text{FDR}} = 0.407$). Deservingness/Worthlessness had insufficient discordant pairs for a valid McNemar's test due to its low base rate (0.5\% in first posts, 0.8\% in last posts). These results converge with the GEE findings in supporting overall stability of self-stigma expression patterns.

\begin{table}[ht]
\centering
\caption{Indicator Prevalence in First Versus Last Post ($n$ = 384 users)}
\label{tab:supp_first_last}
\small
\begin{tabular}{lccccc}
\hline
\textbf{Indicator} & \textbf{First} & \textbf{Last} & \textbf{Change} & \textbf{$p$} & \textbf{$p_{\text{FDR}}$} \\
\hline
Self-Labeling & 16.1\% & 12.0\% & $-$4.2 pp & 0.094 & 0.210 \\
Pessimism/Self-Defeatism & 6.5\% & 3.4\% & $-$3.1 pp & 0.067 & 0.200 \\
Deservingness/Worthlessness & 0.5\% & 0.8\% & +0.3 pp & --- & --- \\
Shame & 2.9\% & 2.9\% & 0.0 pp & 0.831 & 0.831 \\
Guilt/Self-Blame & 6.0\% & 2.6\% & $-$3.4 pp & 0.021 & 0.188 \\
Despair/Hopelessness & 10.9\% & 10.2\% & $-$0.8 pp & 0.810 & 0.831 \\
Concealment & 6.3\% & 4.2\% & $-$2.1 pp & 0.256 & 0.407 \\
Anticipated Rejection & 8.1\% & 6.5\% & $-$1.6 pp & 0.470 & 0.605 \\
Desire to Quit & 18.5\% & 21.9\% & +3.4 pp & 0.271 & 0.407 \\
Ambivalence & 13.5\% & 9.1\% & $-$4.4 pp & 0.054 & 0.200 \\
\hline
\multicolumn{6}{p{11cm}}{\footnotesize\textit{Note:} pp = percentage points. $p$ from McNemar's test. $p_{\text{FDR}}$ = Benjamini-Hochberg adjusted. --- indicates insufficient cell counts for valid test. No indicators significant at $p_{\text{FDR}} < 0.05$.} \\
\end{tabular}
\end{table}

\subsection{Individual Indicator Trends}

To assess whether individual indicators showed systematic temporal change, we fit separate binary GEE models (logit link, exchangeable working correlation) for each of the ten indicators, with normalized post position as the predictor and the binary indicator label as the outcome. These models were restricted to self-stigma posts ($N = 3{,}838$ posts from 384 users) to examine how the composition of self-stigma expression changed across users' posting trajectories, conditional on expressing self-stigma. P-values were adjusted using the Benjamini--Hochberg false discovery rate correction across the ten tests.

Table~\ref{tab:supp_gee_temporal} presents the results. Nine of ten indicators showed no significant temporal change after FDR correction. The sole exception was Pessimism/Self-Defeatism, which showed a significant increase across users' posting trajectories (OR = 1.62, 95\% CI [1.25, 2.10], $p_{\text{FDR}} = 0.002$), indicating that the odds of expressing pessimistic or self-defeatist beliefs were approximately 62\% higher in users' later self-stigma posts compared to their earlier ones. Ambivalence showed a nominal decrease (OR = 0.78, 95\% CI [0.61, 0.99], $p = 0.038$) that did not survive FDR correction ($p_{\text{FDR}} = 0.189$). The divergent trajectories of pessimism (increasing) and ambivalence (directionally decreasing) are conceptually consistent with a shift from active motivational conflict toward settled resignation, though the ambivalence trend should be interpreted cautiously given its non-significance after correction.

\begin{table*}[ht]
\centering
\caption{Temporal Change in Indicator Prevalence (GEE Models)}
\label{tab:supp_gee_temporal}
\small
\begin{tabular}{llcccccc}
\hline
\textbf{Domain} & \textbf{Indicator} & \textbf{$\beta$} & \textbf{SE} & \textbf{OR} & \textbf{95\% CI} & \textbf{$p$} & \textbf{$p_{\text{FDR}}$} \\
\hline
Cognitive & Self-Labeling & 0.05 & 0.12 & 1.06 & [0.83, 1.34] & 0.658 & 0.743 \\
Cognitive & Pessimism/Self-Defeatism & 0.49 & 0.13 & 1.62 & [1.25, 2.10] & $<$0.001 & 0.002** \\
Cognitive & Deservingness/Worthlessness & 0.31 & 0.23 & 1.36 & [0.87, 2.13] & 0.171 & 0.426 \\
Affective & Shame & 0.07 & 0.20 & 1.07 & [0.73, 1.57] & 0.739 & 0.743 \\
Affective & Guilt/Self-Blame & $-$0.13 & 0.16 & 0.88 & [0.64, 1.21] & 0.439 & 0.627 \\
Affective & Despair/Hopelessness & 0.14 & 0.12 & 1.15 & [0.91, 1.45] & 0.257 & 0.514 \\
Behavioral & Concealment & $-$0.19 & 0.19 & 0.83 & [0.58, 1.19] & 0.316 & 0.527 \\
Behavioral & Anticipated Rejection & 0.05 & 0.15 & 1.05 & [0.79, 1.40] & 0.743 & 0.743 \\
Behavioral & Desire to Quit & 0.21 & 0.14 & 1.23 & [0.95, 1.61] & 0.121 & 0.404 \\
Behavioral & Ambivalence & $-$0.25 & 0.12 & 0.78 & [0.61, 0.99] & 0.038 & 0.189 \\
\hline
\multicolumn{8}{p{14cm}}{\footnotesize\textit{Note:} GEE models with exchangeable correlation, restricted to self-stigma posts ($N$ = 3,838 posts from 384 users). OR = odds ratio for one-unit increase in normalized post position (0 = first SS post, 1 = last SS post). **$p_{\text{FDR}} < 0.01$.} \\
\end{tabular}
\end{table*}

\subsection{Episode-Level Sensitivity Analysis}\label{supp:robustness}

The primary transition analysis treated each self-stigma post as a separate observation. To test whether results were sensitive to this choice, we repeated the state classification and transition analysis after merging temporally adjacent self-stigma posts into episodes using three window sizes: 24-hour, 48-hour, and 7-day. Within each episode, we took the union of all indicator labels, then assigned the episode to a compositional state using the same hierarchy described in Section 2.4.2 of the main text.

Table~\ref{tab:supp_robustness_episodes} presents the results. Merging posts into episodes compressed the dataset by 18.6\% (24-hour window) to 30.1\% (7-day window), reducing the number of users meeting the 3+ episode threshold from 384 to 318--347, respectively. Despite this compression, the key qualitative finding was preserved across all windows: Affective-present showed the highest self-persistence rate (67.4--69.8\%), substantially exceeding the self-persistence rates of all other states (8.0--18.3\%). The rank ordering of state persistence rates was unchanged, and the dominance of Affective-present as the most common destination from every non-affective state was maintained at all window sizes.

\begin{table}[h]
\centering
\caption{State Persistence Rates Across Episode-Merging Windows}
\label{tab:supp_robustness_episodes}
\begin{tabular}{lcccc}
\toprule
 & \multicolumn{4}{c}{\textbf{Self-Persistence Rate (\%)}} \\
\cmidrule(lr){2-5}
\textbf{Window} & Behav.-pred. & Cog.-only & Cog.-Behav. & Aff.-present \\
\midrule
None (primary) & 25.1 & 20.8 & 29.4 & 72.2 \\
24-hour & 12.3 & 8.0 & 18.3 & 67.6 \\
48-hour & 12.2 & 8.2 & 18.1 & 67.4 \\
7-day & 13.6 & 9.1 & 17.4 & 69.8 \\
\bottomrule
\end{tabular}

\vspace{4pt}
\parbox{0.9\textwidth}{\footnotesize\textit{Note:} Self-persistence rate = probability that a post/episode in a given state is followed by a post/episode in the same state. ``None'' row reproduces the diagonal of Table 7 of the main text for comparison. Episode counts: 24-hour = 2,207 episodes (18.6\% compression), 48-hour = 2,136 (21.2\%), 7-day = 1,894 (30.1\%). Users with 3+ episodes: 347, 343, and 318, respectively.}
\end{table}

The reduction in self-persistence rates for non-affective states under episode merging is expected: merging temporally adjacent posts increases the probability that at least one post in the episode contains an affective indicator, reclassifying what would have been a non-affective state into Affective-present. That affective persistence remained high (67--70\%) even after this reclassification underscores the robustness of the finding.

\paragraph{Posting Tempo and Next-Post Composition.}
We also examined whether the time gap between consecutive self-stigma posts affected the composition of the subsequent post, using GEE models to account for within-user clustering. The log-transformed gap between consecutive self-stigma posts did not predict the number of indicators in the next post ($\beta = 0.003$, SE $= 0.019$, $p = 0.87$), indicating that posting tempo was unrelated to expression complexity. However, longer gaps were associated with modestly lower probability that the next post would be classified as Affective-present (OR $= 0.941$, 95\% CI [0.903, 0.982], $p = 0.005$). Descriptively, 64\% of posts following a gap of $\leq$7 days were Affective-present, compared to 57\% following gaps of $>$30 days. This pattern may reflect that users returning after longer absences are more likely to re-enter with action-oriented or cognitive framing (e.g., desire to quit, self-labeling) rather than immediately disclosing affective experiences, consistent with the behavioral-first disclosure pattern observed in the emergence analysis (Section 2.4.2 of the main text).

\end{document}